\documentclass[lettersize,journal]{IEEEtran}
\usepackage{amsmath,amsfonts}
\usepackage{algorithmic}
\usepackage{algorithm}
\usepackage{array}
\usepackage[caption=false,font=normalsize,labelfont=sf,textfont=sf]{subfig}
\usepackage{textcomp}
\usepackage{stfloats}
\usepackage{url}
\usepackage{verbatim}
\usepackage{graphicx}
\usepackage{cite}
\usepackage{bm}
\usepackage{multicol}
\usepackage{multirow}
\usepackage{float}
\usepackage{makecell}
\usepackage{tabularx}
\usepackage{booktabs}
\usepackage{subcaption}
\usepackage[square,sort,comma,numbers]{natbib}
\begin{document}

\title{Beyond Semantics: An Evidential Reasoning-Aware Multi-View Learning Framework for Trustworthy Mental Health Prediction}




\author{Yucheng Ruan\IEEEauthorrefmark{1},
Ling Huang\IEEEauthorrefmark{2}, 
Qika Lin\IEEEauthorrefmark{1},
Kai He\IEEEauthorrefmark{1},
Mengling Feng\IEEEauthorrefmark{1}
\thanks{\IEEEauthorrefmark{1}
Yucheng Ruan, Qika Lin, Kai He, Mengling Feng are with Saw Swee Hock School of Public Health,
National University of Singapore, Singapore.
}
\thanks{\IEEEauthorrefmark{2}
Ling Huang (corresponding author) is with the Clinical Science, Imperial College London, United Kingdom.
Email: my@linghuang.co.uk}
}

\markboth{IEEE Transactions on Affective Computing}%
{Shell \MakeLowercase{\textit{et al.}}: A Sample Article Using IEEEtran.cls for IEEE Journals}


\maketitle

\begin{abstract}
Automated mental health prediction using textual data has shown promising results with deep learning and large language models. However, deploying these models in high-stakes real-world settings remains challenging, as existing approaches largely rely on semantic representations and often produce overconfident predictions under ambiguous, noisy, or shifted data. Moreover, most methods lack reliable uncertainty estimation, undermining trust in risk-sensitive mental health applications.
To address these limitations, we formulate the task as a multi-view learning problem that integrates semantic information from encoder-only models with higher-level reasoning information from decoder-only models, where reasoning-aware representations and uncertainty modeling are obtained in a trustworthy manner.  
To ensure reliable fusion, we adopt an evidential learning framework based on Subjective Logic to explicitly model uncertainty and introduce an evidential fusion strategy that balances complementary views while discounting unreliable evidence. 
Benchmarking on three real-world datasets, Dreaddit, SDCNL, and DepSeverity, reports accuracies of 0.835, 0.731, and 0.751, respectively, demonstrating its potential for reliable mental health prediction. Additional experiments on robustness to noise and case studies for interpretability confirm that our proposed framework not only improves predictive performance but also provides trustworthy uncertainty estimates and human-understandable reasoning signals, making it suitable for risk-sensitive applications in mental health assessment.
\end{abstract}


\section{Introduction}

\begin{figure}
    \centering
    \includegraphics[width=1\linewidth]{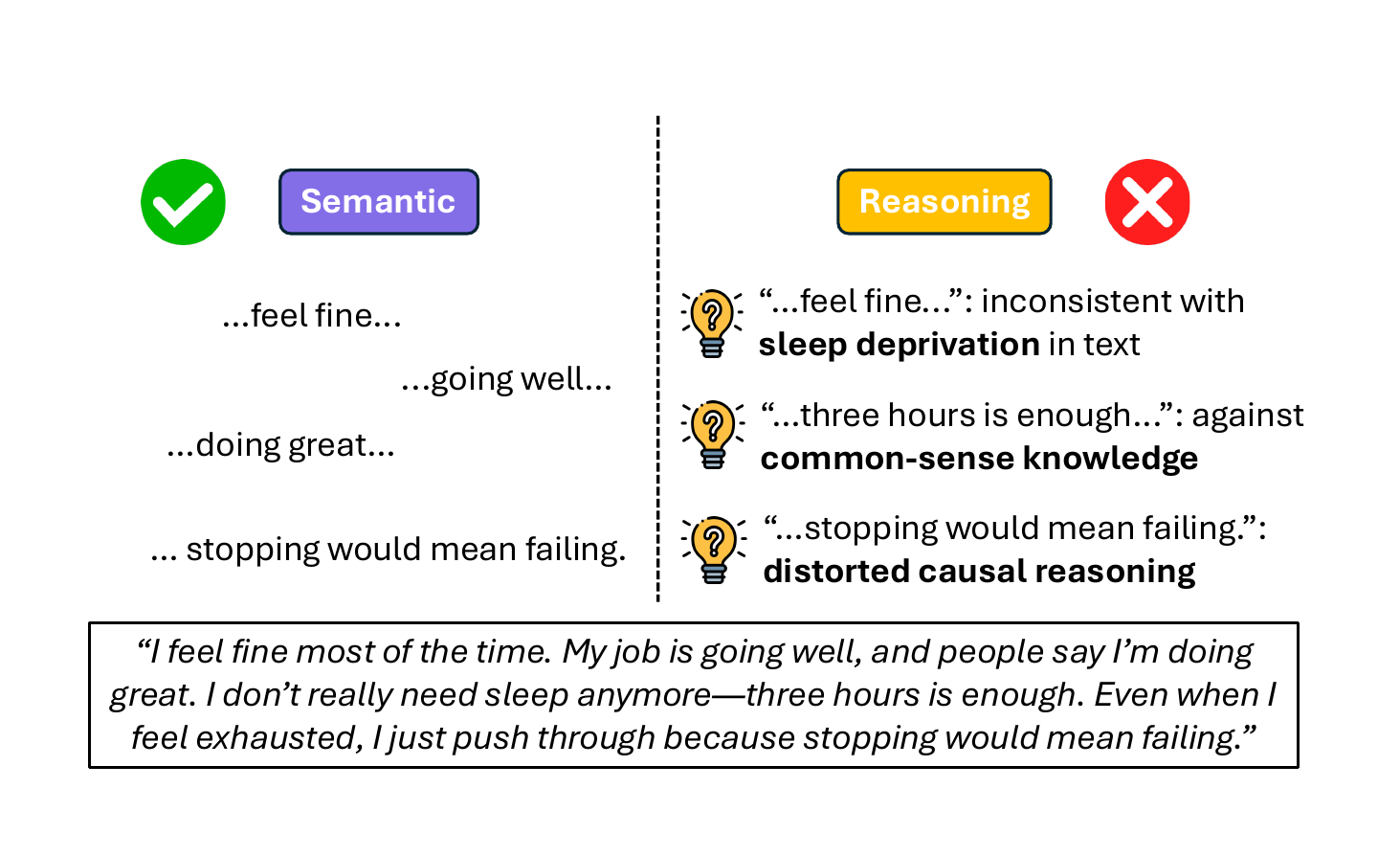}
    \caption{Analysis on mental health narrative from both semantic and reasoning perspectives. From semantic perspective, the text appears predominantly positive, containing expressions such as “fine,” “going well,” and “doing great.” while the reasoning-oriented analysis reveals underlying conflicts, including logical inconsistencies, maladaptive causal reasoning, and violations of basic physiological norms.}
    \label{fig:sample}
\end{figure}

Mental health disorders have become a major global public health concern, driving growing interest in automated mental health prediction using machine learning models \cite{world2017depression}. With the rapid development of deep learning and large language models (LLMs), recent studies have reported promising results by leveraging textual data such as social media posts, online forums, and self-reported narratives \cite{jin2025applications,abd2020application}. Conventional approaches mainly rely on semantic representations learned by encoder-only models, such as BERT \cite{devlin2019bert}, to capture linguistic meaning and support large-scale mental health detection \cite{ge2025survey}. Although these methods have shown strong empirical performance, their deployment in real-world mental health settings remains challenging due to the high-stakes nature of such applications. In particular, they often do not explicitly model higher-level reasoning signals, including causal relations, emotional dynamics, or internal inconsistencies that frequently appear in mental health narratives, as illustrated in Figure~\ref{fig:sample}. As a result, these models may produce confident predictions even when the supporting evidence is weak, ambiguous, or conflicting.

Recent studies \cite{lamichhane2023evaluation,yang2023evaluations,xu2024mental} have shown that generative LLMs exhibit preliminary capabilities in predicting mental health conditions through zero-shot or few-shot learning, suggesting their potential for reasoning over complex texts. However, their performance remains limited compared to state-of-the-art domain-specific models, and their reliability in high-risk scenarios is not well studied \cite{huang2024trustllm}.

Moreover, human mental health assessment processes are rarely based on isolated semantic cues; instead, they involve reasoning over multiple pieces of evidence, and evaluating the coherence of expressed thoughts and emotions \cite{thagard2018mental}. Inspired by this process, we formulate mental health prediction as a multi-view learning problem, where semantic and reasoning views provide complementary information. 
Specifically, the semantic view, derived from encoder-only models, captures the overall semantic content and linguistic style of the text, while the reasoning view, obtained from decoder-only models, also known as generative models, focuses on higher-level language reasoning information, such as emotional patterns and cognitive states reflected in the narrative.

However, directly fusing information from heterogeneous representations may potentially harm prediction reliability~\cite{wan2024knowledge}. 
This challenge is further amplified in realistic scenarios involving noisy data, ambiguous expressions, or distribution shifts across populations \cite{ovadia2019can,zhang2024multimodal,keluskar2024llms}. In such settings, models must provide reliable uncertainty estimates to support risk-sensitive decision making. 
Existing uncertainty-aware approaches, such as Bayesian neural networks and ensemble methods, attempt to quantify predictive uncertainty \cite{geifman2017selective,smith2018understanding,lakshminarayanan2017simple}, but often suffer from poor calibration and weak correlation with actual misclassification risk, particularly under data noise \cite{ovadia2019can}. Moreover, most of these methods treat uncertainty as a byproduct of prediction, without explicitly considering whether the model’s internal reasoning process is reliable.

To address these challenges, we propose a reasoning-aware multi-view learning framework with evidential fusion for trustworthy mental health prediction. Beyond extracting semantic representations with an encoder-only language model, our approach learns high-level reasoning representations using decoder-only models. It models predictions as Dirichlet distribution via Subjective Logic \cite{josang2016subjective}, allowing explicit quantification of both predictive confidence and uncertainty. An evidential fusion strategy then integrates semantic and reasoning views while accounting for their respective uncertainties. Experiments on benchmark mental health datasets demonstrate that this framework not only achieves competitive predictive performance but also improves uncertainty estimation, showing stronger alignment between predicted uncertainty and actual errors and greater robustness to textual noise.

In summary, our main contributions are as follows:
\begin{itemize}
\item We propose a reasoning-aware, evidential multi-view learning framework for mental health prediction that extends beyond semantic representations to incorporate high-level reasoning information.
\item We adopt an evidential learning framework in which model uncertainty is explicitly represented using Subjective Logic, and complementary views are integrated through an evidential fusion strategy to provide reliable uncertainty estimates.
\item We conduct extensive experiments on three mental health-related datasets to assess predictive performance and uncertainty estimation. Our framework consistently outperforms existing baselines across multiple datasets. Moreover, it remains robust under realistic typographical noise and provides interpretable predictions, highlighting its practical value for mental health applications.
\end{itemize}

\section{Related Work}
\subsection{Language Models for Mental Health Prediction}
Language models, from early architectures such as BERT \cite{devlin2019bert} and GPT \cite{radford2018improving} to instruction-tuned larger models like GPT-4 \cite{achiam2023gpt} and LLaMA \cite{touvron2023llama}, have shown strong generalization across diverse natural language tasks. Recent studies have explored their use in mental health prediction, including detecting depression, anxiety, stress, and suicidal ideation from social media posts and self-reported narratives \cite{ma2024integrating}. While general-purpose LLMs show promising zero-shot and few-shot performance, they often underperform task-specific models due to challenges in domain alignment and robustness \cite{kim2024large}. Domain-adaptive fine-tuning with mental health–specific data has been shown to improve predictive performance \cite{xu2024mental}. Despite these advances, research on reliable and trustworthy LLM-based mental health prediction remains limited, highlighting the need for methods that can handle high-stakes decision-making.

\subsection{Multi-View Learning}
Multi-view learning leverages multiple complementary representations of data to improve robustness and performance. Traditional approaches focus on learning shared and view-specific representations or aligning views to capture complementary information~\cite{zhang2019cpm,chen2020simple}. Recent surveys indicate that deep neural network–based multi-view models have advanced classification, clustering, and representation fusion by improving nonlinear modeling and scalability~\cite{xu2013survey,yu2025review}. Most existing methods assume views come from fixed modalities or augmentations and focus on alignment or joint objectives. They do not, however, exploit the semantic and reasoning diversity that can be induced by LLMs, where each view can capture distinct perspectives beyond simple feature variations.

\subsection{Uncertainty-Aware Learning}
Reliable uncertainty estimation is crucial for safety-sensitive applications. Traditional approaches, including Monte Carlo dropout \cite{gal2016dropout} and deep ensembles \cite{lakshminarayanan2017simple}, require additional inference computations and often struggle to jointly optimize accuracy, robustness, and uncertainty. Evidential Deep Learning (EDL) \cite{sensoy2018evidential} addresses this by parameterizing uncertainty using belief and uncertainty masses without sampling. Extensions of EDL to multi-view settings allow uncertainty-aware fusion, down-weighting less reliable views during decision-making \cite{han2021trustedmultiviewclassification,han2022trusted}. However, existing techniques are typically designed for fixed views and do not provide practical insights for mental health prediction in real-world, high-uncertainty scenarios. In contrast, our approach (1) leverages LLMs to generate semantic and reasoning views, (2) incorporates an uncertainty estimation framework for LLM-based mental health prediction, and (3) provides a detailed analysis of prediction reliability in this domain.

\section{Preliminaries}
\subsection{Evidence theory}
Evidence theory, commonly referred to as Dempster-Shafer theory, provides a general mathematical framework for modeling uncertainty when information is incomplete, imprecise, or conflicting \cite{dempster1967upper,shafer1976mathematical}. A key distinction from classical probability theory is that belief can be assigned not only to individual hypotheses but also to their subsets, allowing explicit representation of ignorance.

Let $\Omega = \{\omega_1, \omega_2, \cdots, \omega_K\}$ denote a finite \emph{frame of discernment}. A body of evidence is encoded by a basic probability assignment (mass function) $m: 2^{\Omega} \rightarrow [0,1]$, satisfying $m(\emptyset)=0$ and
\begin{equation}
\sum_{A \subseteq \Omega} m(A) = 1.
\end{equation}
For any subset $A \subseteq \Omega$, the value $m(A)$ represents the belief that supports exactly the proposition that the true hypothesis lies in $A$, without committing belief to any of its strict subsets. Subsets with positive mass are referred to as \emph{focal elements}. When belief is assigned exclusively to singleton sets, the mass function coincides with a Bayesian probability distribution.

In multi-class classification, it is common to restrict focal elements to the $K$ singleton hypotheses and the full set $\Omega$, corresponding to class-specific belief and residual uncertainty, respectively. Under this setting, the mass function can be written compactly as
\begin{equation}
    m = \big(m(\{\omega_1\}), m(\{\omega_2\}), \cdots, m(\{\omega_K\}), m(\Omega)\big) \in \mathbb{R}^{K+1},
\end{equation}
where $m(\{\omega_k\})$ quantifies the belief in class $\omega_k$, and $m(\Omega)$ captures the remaining uncertainty due to insufficient or ambiguous evidence. 
Based on this formulation, evidential neural networks (ENNs) have been developed to map uncertain inputs into belief mass assignments rather than point estimates, enabling uncertainty-aware predictions suitable for trusted decision-making \cite{denoeux2000neural,huang2024review}. Beyond ENNs, Subjective Logic (SL) has been applied in medical imaging tasks \cite{zou2025toward,wang2025uncertainty,zhou2025medsam}, offering a more granular representation of uncertainty by separately modeling belief, disbelief, and ignorance in a Dirichlet distribution. It can be implemented with fewer parameters, reducing computational overhead while enhancing interpretability and robustness of predictions.

\subsection{Dempster’s Rule of Combination}

Given two independent sources of evidence represented by mass vectors
\[
m^{(1)} = \big(m^{(1)}(\{\omega_1\}), \dots, m^{(1)}(\{\omega_M\}), m^{(1)}(\Omega)\big)
\]
and
\[
m^{(2)} = \big(m^{(2)}(\{\omega_1\}), \dots, m^{(2)}(\{\omega_M\}), m^{(2)}(\Omega)\big),
\]
Dempster’s rule provides a principled mechanism for aggregating them into a joint mass function
\begin{equation}
m = m^{(1)} \oplus m^{(2)}.
\end{equation}

Under the singleton--$\Omega$ assumption, the combined belief assigned to class $\omega_k$ and the combined uncertainty are given by
\begin{align}
m(\{\omega_k\}) &= \frac{1}{1-\kappa}\Big(m^{(1)}(\{\omega_k\}) m^{(2)}(\{\omega_k\}) \\
&+ m^{(1)}(\{\omega_k\}) m^{(2)}(\Omega)
+ m^{(2)}(\{\omega_k\}) m^{(1)}(\Omega)
\Big), \\
m(\Omega) &= \frac{1}{1-\kappa} \,
m^{(1)}(\Omega) m^{(2)}(\Omega),
\label{eq:dempster}
\end{align}
where
\begin{equation}
\kappa = \sum_{i \neq j}
m^{(1)}(\{\omega_i\}) m^{(2)}(\{\omega_j\})
\end{equation}
measures the degree of conflict between the two evidence sources. The normalization factor $(1-\kappa)^{-1}$ ensures that the resulting mass function remains valid.

\begin{figure*}
    \centering
    \includegraphics[width=1\linewidth]{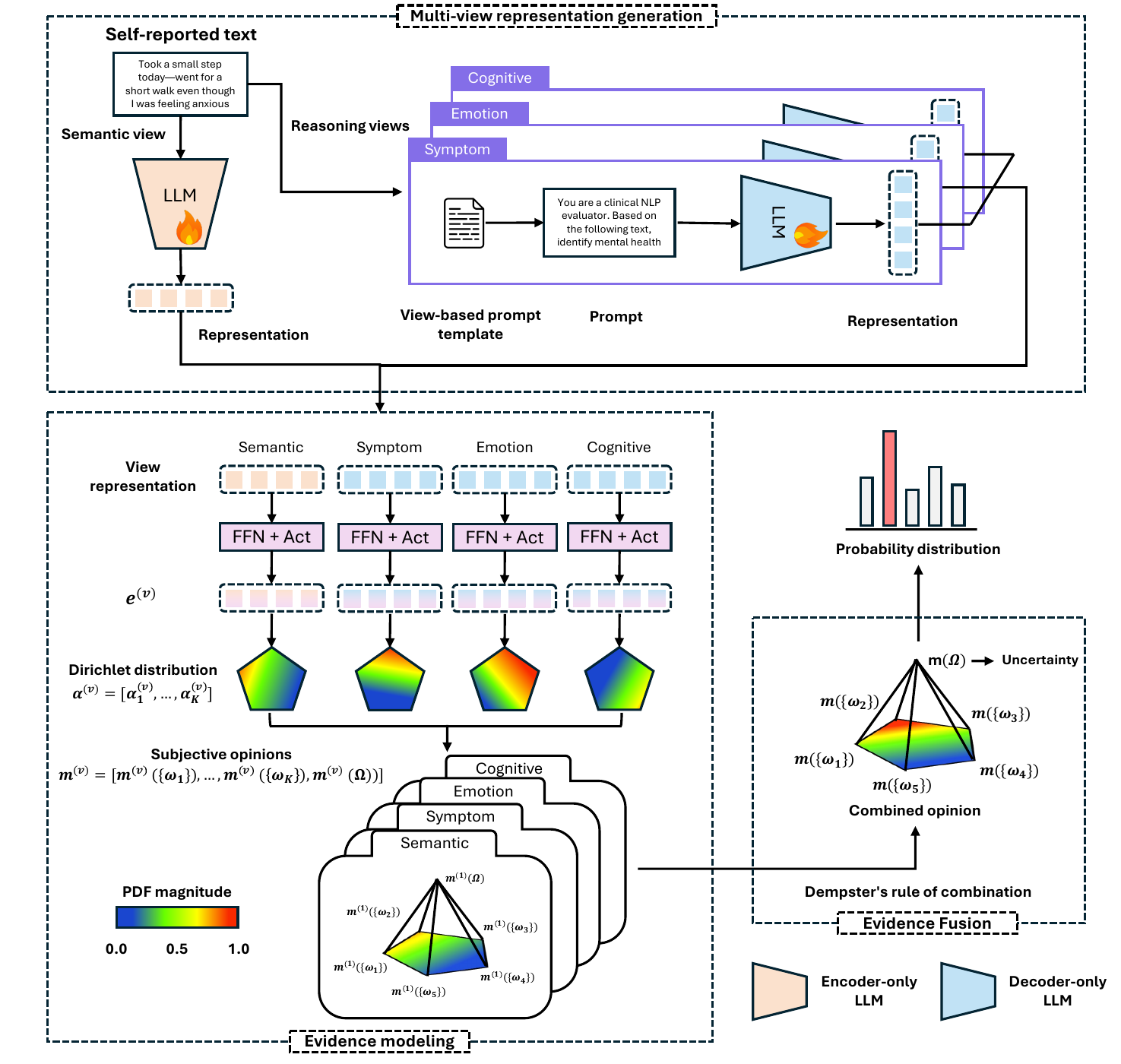}
    \caption{Overview of the proposed model, which consists of three key stages: (1) multi-view representation generation with LLM, (2) evidence modeling with subjective logic, and (3) multi-view evidence fusion.}
    \label{fig:overview}
\end{figure*}
\section{Methodology}
In this section, we present our methodology, which consists of three stages: multi-view representation generation using a large language model, evidence modeling based on subjective logic, and multi-view evidence fusion, as illustrated in Figure~\ref{fig:overview}.
\subsection{Multi-View Representation Generation}
In this section, we introduce the representation generation with different LLMs from different views. This multi-view approach is designed to capture both the dense, contextual semantics and rich, perspective-driven reasoning inherent in the data. The semantic view provides a stable, distributional understanding of the input, while the reasoning view offers a flexible, interpretable analysis from multiple conceptual angles. This hybrid setup is posited to offer a more robust foundation for the subsequent evidential fusion and final classification than any single-view approach.

\subsubsection{Semantic View}
The semantic view is implemented using a pre-trained encoder architecture, which can encapsulate the overall semantic content, sentiment, and linguistic style of the input text \cite{ji2022mentalbert}, and we adopt \textit{BERT} in our study. For the downstream classification task, the final hidden state associated with the $[CLS]$ token is extracted as the representative feature vector to aggregate global sequence information into a single, high-dimensional representation suitable for mental health prediction.

\subsubsection{Reasoning Views}
In contrast, the reasoning views are generated using a decoder-only autoregressive language model (we adopt \textit{LLAMA-3-8B-Instruct} in our study). This process leverages the theory of prompt learning, where the original user text is analyzed under $T$ ($T$ is the number of reasoning views) distinct instructional prompts (e.g., a clinical diagnostic prompt, a cognitive-behavioral prompt). For each prompt, the model autoregressively generates a coherent chain-of-thought analysis. To convert this free-text reasoning into a fixed-dimensional feature vector, the hidden state of the last token in the generated sequence is utilized. In autoregressive decoder models, the hidden state at each step accumulates context from all previous tokens; therefore, the final token's state serves as a comprehensive representation of the entire generated analysis. Each of the $T$ reasoning pathways thus produces a unique feature vector, collectively forming a multi-faceted Reasoning View that captures diverse interpretive dimensions of the input text.

Mental health conditions manifest through interconnected symptomatic, emotional, and cognitive channels. To model this complexity, our reasoning framework deconstructs the textual input into three specialized analytical perspectives. This tri-view approach moves beyond unitary classification, enabling the model to mimic a holistic clinical assessment by evaluating \textbf{what} a person experiences (Symptom View), \textbf{how} they feel (Emotion View), and \textbf{how} they think (Cognitive View).

\paragraph{Symptom View}
The Symptom View is designed to extract explicit and implicit indicators of psychological distress that align with established clinical criteria (e.g., DSM-5 \cite{regier2013dsm}). In mental health diagnostics, symptom reporting is foundational. This view operationalizes this by prompting the LLM to act as a clinical observer, scanning the text for descriptions of behaviors, somatic complaints, and self-reported states. The prompt instructs the model to generate concise, discrete symptom phrases. For example, from a narrative about daily life, it might extract outputs such as “persistent low energy,” “disrupted sleep patterns,” or “loss of appetite.” The resulting representation, derived from the final token's hidden state of this symptom-focused generation, provides a structured feature set that grounds the classification in phenomenology, directly connecting language to clinically-relevant constructs.

\paragraph{Emotion View}
The Emotion View captures the affective substrate of the text, which is often a more immediate and nuanced indicator of mental state than symptom reporting alone \cite{lalk2025employing}. Emotional tone and volatility are critical in assessing conditions like depression and anxiety \cite{schlicher2025emotionally}. This view prompts the LLM to perform a fine-grained emotional analysis, identifying dominant emotions, their intensity, and shifts within the narrative. It focuses on the person's expressed feelings, such as “overwhelming guilt,” “pervasive anxiety,” or “emotional numbness.” By converting this affective analysis into a feature vector, the model encodes the emotional valence and arousal present in the text. This view complements the Symptom View by adding a layer of subjective feeling, helping to distinguish between, for instance, reported fatigue with accompanying distress versus fatigue without affective complaint.

\paragraph{Cognitive View}
The Cognitive View targets the underlying thought patterns that can maintain or exacerbate mental health conditions, a cornerstone of Cognitive Behavioral Therapy (CBT) \cite{nakao2021cognitive}. This view prompts the LLM to identify and label cognitive distortions and maladaptive thinking styles present in the text. It analyzes the logic, perspective, and narrative style, generating outputs such as “catastrophizing (‘This minor mistake means I will be fired’),” “dichotomous thinking (‘I am either a success or a total failure’),” or “overgeneralization (‘Nothing ever goes right for me’).” The representation from this view encapsulates the structural quality of thinking. This is crucial for prediction, as specific cognitive patterns are strongly linked to particular disorders, offering explanatory power and differentiating between conditions with overlapping symptomatic presentations.

By incorporating the symptom, emotion, and cognitive views, our approach produces a comprehensive, multi-dimensional reasoning representation. This threefold structure ensures that the evidence supporting the model’s final classification covers key aspects of human psychological experience, thereby improving both the robustness and clinical plausibility of predictions compared to models that rely on a single semantic perspective. The prompt templates used for each view are provided in Table \ref{tab:prompt_templates}.

\begin{table*}[t]
\centering
\caption{Prompt templates for symptom, emotion, and cognitive views.}
\label{tab:prompt_templates}
\begin{tabular}{|p{0.15\textwidth}|p{0.78\textwidth}|}
\hline
\textbf{View} & \textbf{Prompt Template} \\
\hline
Symptom View &
\begin{minipage}[t]{\linewidth}
You are a clinical NLP evaluator. Based on the following text, identify mental health symptoms the person is expressing.\\[2mm]
\textbf{Text:} \{text\} \\[1mm]
List potential symptoms, referencing psychological terminology (e.g., hopelessness, loss of interest, sleep disturbance, guilt, anxiety, trauma reactions).\\[1mm]
\textbf{Output:}
\begin{itemize}
    \item Symptoms detected:
    \item Evidence sentences:
    \item Psychological interpretation:
\end{itemize}
\end{minipage} \\
\hline
Emotion View &
\begin{minipage}[t]{\linewidth}
You are a psychological expert in emotional text analysis. Analyze the emotional content of the text below.\\[2mm]
\textbf{Text:} \{text\} \\[1mm]
\textbf{Output:}
\begin{itemize}
    \item Emotion distribution (0--1 scale): sadness, anxiety, anger, fear, joy, numbness, loneliness.
    \item Primary emotions expressed.
    \item Emotional intensity level (low / medium / high).
    \item Is there emotional instability or rapid shift? Provide evidence.
\end{itemize}
\end{minipage} \\
\hline
Cognitive View &
\begin{minipage}[t]{\linewidth}
You are a cognitive behavioral therapist. Identify cognitive distortions present in the text.\\[2mm]
\textbf{Text:} \{text\} \\[1mm]
Check for:
\begin{itemize}
    \item Catastrophizing
    \item Overgeneralization
    \item Black-and-white thinking
    \item Personalization
    \item Mind-reading assumptions
    \item Hopeless reasoning
    \item Rumination
\end{itemize}
\textbf{Output:}
\begin{itemize}
    \item Distortions detected:
    \item Sentences or phrases showing them:
    \item CBT-based interpretation:
\end{itemize}
\end{minipage} \\
\hline
\end{tabular}

\end{table*}

\subsection{Evidence Modeling}

Here, we reformulate \emph{Subjective Logic} (SL) \cite{jsang2018subjective} for uncertainty-aware evidence modeling.
SL provides a principled connection between observed evidence and probabilistic belief by interpreting predictions through a Dirichlet distribution, which serves as the conjugate prior of the categorical distribution \cite{bishop2006pattern}. This formulation allows uncertainty to be explicitly quantified and directly regulated by the amount of supporting evidence.

In contrast to prototype-based evidential networks, where uncertainty arises implicitly from geometric relations in the embedding space and often incurs additional computational complexity, SL-based models yield semantically interpretable and theoretically grounded uncertainty estimates. Moreover, the explicit belief–uncertainty decomposition in Subjective Logic naturally supports evidence fusion across multiple views, making it well-suited for trusted decision-making.

For a $K$-class classification problem and the $v^{\text{th}}$ view, Subjective Logic represents predictive uncertainty using a belief mass assigned to each class and a residual mass corresponding to overall uncertainty. These masses satisfy the normalization constraint
\begin{equation}
m^{(v)}(\Omega) + \sum_{k=1}^{K} m^{(v)}(\{\omega_k\}) = 1,
\label{eq:sl_norm}
\end{equation}
where $m^{(v)}(\{\omega_k\}) \geq 0$ denotes the belief supporting class $\omega_k$, and $m^{(v)}(\Omega) \geq 0$ captures the remaining uncertainty caused by insufficient or ambiguous evidence.

Firstly, the final softmax layer is replaced by a non-negative activation function (e.g. Softplus \cite{dugas2000incorporating}), ensuring that the network outputs a non-negative evidence vector 
\begin{equation}
    \mathbf{e}^{(v)} = [e^{(v)}_1, \dots, e^{(v)}_K].
\end{equation}
To construct the belief masses, SL links the non-negative evidence vector to the parameters of a Dirichlet distribution

\begin{equation}
    \boldsymbol{\alpha}^{(v)} = [\alpha^{(v)}_1, \dots, \alpha^{(v)}_K],
\end{equation}
where each concentration parameter is defined as
\begin{equation}
\alpha_k^{(v)} = e_k^{(v)} + 1.
\end{equation}
Let $S^{(v)} = \sum_{i=1}^{K} \alpha_i^{(v)}$ denote the Dirichlet strength. The belief mass for class $\omega_k$ and the overall uncertainty mass are then computed as
\begin{equation}
\begin{aligned}
m^{(v)}(\{\omega_k\}) &= \frac{e_k^{(v)}}{S^{(v)}} 
= \frac{\alpha_k^{(v)} - 1}{S^{(v)}}, \\
m^{(v)}(\Omega) &= \frac{K}{S^{(v)}}.
\end{aligned}
\label{eq:sl_mass}
\end{equation}

This formulation reflects the intuitive behavior that increasing evidence for a specific class strengthens its associated belief, while limited total evidence leads to higher predictive uncertainty. The resulting mass assignment constitutes a \emph{subjective opinion} over the frame of discernment. Given this opinion, the expected class probability induced by the corresponding Dirichlet distribution is obtained as
\begin{equation}
\hat{p}_k^{(v)} = \frac{\alpha_k^{(v)}}{S^{(v)}}.
\end{equation}

Unlike conventional neural classifiers that output a single probability vector lying on the simplex, Subjective Logic models a distribution over such probability assignments. As a result, it captures second-order uncertainty, providing a richer and more reliable characterization of predictive confidence.


\subsection{Multi-View Evidence Fusion}
With belief and uncertainty obtained for each individual view, Dempster-Shafer theory provides a principled mechanism for aggregating evidence from multiple independent sources into a single, coherent belief assignment. In particular, given $V$ views, each view $v$ yields a mass function
\[
m^{(v)} = \big(m^{(v)}(\{\omega_1\}), \dots, m^{(v)}(\{\omega_K\}), m^{(v)}(\Omega)\big),
\]
which encodes class-wise belief and residual uncertainty based on the corresponding evidence.

To integrate information across views, the mass functions are assumed to be mutually independent and are iteratively combined using Dempster’s rule of combination. The resulting joint mass function is given by
\begin{equation}
m = m^{(1)} \oplus m^{(2)} \oplus \cdots \oplus m^{(V)},
\label{eq:fusion2}
\end{equation}
where $\oplus$ denotes the Dempster--Shafer fusion operator. This operation jointly aggregates class-specific belief masses and uncertainty masses while accounting for potential conflicts among views.

After fusion, the combined mass function
\begin{equation}
    m = \big(m(\{\omega_1\}), \dots, m(\{\omega_K\}), m(\Omega)\big)
\end{equation}
can be mapped back to the Subjective Logic parameterization. Following the inverse relationship between belief, uncertainty, and evidence, the total Dirichlet strength and class-wise evidence are computed as
\begin{equation}
\begin{aligned}
S &= \frac{K}{m(\Omega)}, \\
e_k &= m(\{\omega_k\}) \cdot S, \\
\alpha_k &= e_k + 1,
\end{aligned}
\label{eq:sl2}
\end{equation}
where $S$ denotes the accumulated evidence strength across all views, $e_k$ represents the fused evidence supporting class $\omega_k$, and $\alpha_k$ are the parameters of the resulting Dirichlet distribution.

Through this fusion process, multi-view information is consolidated into a joint evidence vector and a corresponding Dirichlet distribution, from which final class probabilities and an overall measure of predictive uncertainty can be consistently derived.

\subsection{Optimization}
In this section, we describe the optimization process on how neural networks are learned to extract evidence from each view, which is subsequently transformed into view-specific belief mass functions $\{m^{(v)}\}_{v=1}^{V}$ and their aggregated opinion.

Given the evidence for sample $i$, the Dirichlet concentration parameters are defined as $\boldsymbol{\alpha}_i$ which induces a Dirichlet distribution $D(\mathbf{p}_i \mid \boldsymbol{\alpha}_i)$ over the probability simplex, where $\mathbf{p}_i$ represents the latent class probability vector. By taking the expectation of the cross-entropy loss with respect to this Dirichlet distribution, we obtain the adjusted cross-entropy loss (ACE) as below:
\begin{equation}
\begin{aligned}
\mathcal{L}_{\mathrm{ACE}}(\boldsymbol{\alpha}_i)
&= \int\left[\sum_{k=1}^{K}-y_{i k} \log \left(p_{i k}\right)\right] \frac{1}{B\left(\boldsymbol{\alpha}_{i}\right)} \prod_{k=1}^{K} p_{i k}^{\alpha_{i k}-1} d \mathbf{p}_{i} \\
&= \sum_{k=1}^{K} y_{ik}
\big( \psi(S_i) - \psi(\alpha_{ik}) \big),
\end{aligned}
\label{eq:ace}
\end{equation}
where $\psi(\cdot)$ is the digamma function, $B(\bm{\alpha_i})$ denotes the $K$-dimensional multinomial beta function~\cite{johnson1972continuous}, and $p_{i k}$ is the predicted probability of the $i$th sample for class $k$. For our model,and $S_i = \sum_{k=1}^{K} \alpha_{ik}$ denotes the Dirichlet strength. This loss encourages the model to assign larger evidence to the ground-truth class.

However, minimizing $\mathcal{L}_{\mathrm{ACE}}$ alone does not explicitly suppress evidence for incorrect classes. To address this issue, an additional regularization term based on the Kullback--Leibler (KL) divergence is introduced, which penalizes unwarranted evidence by encouraging the Dirichlet distribution to approach a non-informative prior when the prediction is incorrect:
\begin{equation}
\begin{aligned}
KL\big[ D(\mathbf{p}_i \mid &\tilde{\boldsymbol{\alpha}}_i)
\,\|\, D(\mathbf{p}_i \mid \mathbf{1}) \big]
= \log \frac{\Gamma\!\left(\sum_{k=1}^{K} \tilde{\alpha}_{ik}\right)}
{\Gamma(K)\prod_{k=1}^{K}\Gamma(\tilde{\alpha}_{ik})} \\
&\quad + \sum_{k=1}^{K}
(\tilde{\alpha}_{ik}-1)
\Big( \psi(\tilde{\alpha}_{ik})
- \psi\!\left(\sum_{j=1}^{K} \tilde{\alpha}_{ik}\right) \Big),
\end{aligned}
\end{equation}
where $\Gamma(\cdot)$ denotes the gamma function and
\begin{equation}
\tilde{\boldsymbol{\alpha}}_i
= \mathbf{y}_i + ( \mathbf{1} - \mathbf{y}_i ) \odot \boldsymbol{\alpha}_i
\end{equation}
is a modified concentration parameter that avoids penalizing evidence assigned to the ground-truth class to 0.

Combining the two terms, the sample-wise training objective is defined as
\begin{equation}
\mathcal{L}(\boldsymbol{\alpha}_i)
= \mathcal{L}_{\mathrm{ACE}}(\boldsymbol{\alpha}_i)
+ \lambda_t \,
KL\big[ D(\mathbf{p}_i \mid \tilde{\boldsymbol{\alpha}}_i)
\,\|\, D(\mathbf{p}_i \mid \mathbf{1}) \big],
\label{eq:sample_loss}
\end{equation}
where $\lambda_t > 0$ balances classification fidelity and uncertainty regularization. In practice, $\lambda_t$ is gradually increased during training to prevent premature collapse to a uniform distribution.

Finally, to ensure that all views produce coherent and reliable belief mass functions $m^{(v)}$, we adopt a multi-task learning strategy. The overall training objective is given by
\begin{equation}
\begin{aligned}
\mathcal{L}_{\mathrm{overall}}
= \sum_{i=1}^{N}
\left[
\mathcal{L}(\boldsymbol{\alpha}_i)
+ \sum_{v=1}^{V} \mathcal{L}(\boldsymbol{\alpha}_i^{(v)})
\right],
\end{aligned}
\label{eq:overall_loss}
\end{equation}
which jointly optimizes the global prediction and all view-specific evidential models, thereby improving the quality of the resulting fused belief.

\section{Experiments}

\begin{table}[!h]
\caption{Summary of datasets. For each dataset, we reported the task type, number of samples per category, and average text length in tokens.}
\label{tab:dataset}
\begin{tabular}{lllll}
\hline
Dataset                      & \multicolumn{2}{l}{Task Type}             & Data Size & Average Tokens \\ \hline
\multirow{2}{*}{Dreaddit}    & \multirow{2}{*}{Binary}     & Non-stress  & 1696      & 103.2               \\
                             &                             & Stress      & 1857      & 110.5               \\ \hline
\multirow{2}{*}{SDCNL}       & \multirow{2}{*}{Binary}     & Non-Suicide & 915       & 231.0               \\
                             &                             & Suicide     & 980       & 197.3               \\ \hline
\multirow{4}{*}{DepSeverity} & \multirow{4}{*}{Multiclass} & Minimum     & 2587      & 105.3               \\
                             &                             & Mild        & 290       & 114.1               \\
                             &                             & Moderate    & 394       & 111.6               \\
                             &                             & Severe      & 282       & 109.2               \\ \hline
\end{tabular}
\end{table}

\subsection{Datasets}
We conduct the experiments on three real-world datasets:

\subsubsection{Dreaddit \cite{turcan2019dreaddit}}
This dataset was collected retrospectively using the Reddit, covering the period from January 1, 2017 to November 19, 2018. The data consists of user-generated posts from ten subreddits in five domains: abuse, social, anxiety, PTSD, and financial stress. Multiple human annotators assessed whether sentence segments exhibited stress-related signals, and the annotations were aggregated to produce the final labels. We utilized this dataset for post-level binary stress prediction.
It contains 1,696 non-stress and 1,857 stress posts, with an average length of 103.2 and 110.5 tokens per post, respectively.  

\subsubsection{SDCNL \cite{haque2021deep}}
This dataset was collected using the Python Reddit API, including user-generated posts from the r/SuicideWatch and r/Depression subreddits. Each post was manually annotated to indicate the presence of suicidal ideation. We adopted this dataset for post-level binary suicide ideation prediction.
It contains 915 non-suicide and 980 suicide samples, with average text lengths of 231.0 and 197.3 tokens.  

\subsubsection{DepSeverity \cite{naseem2022early}}
This dataset leveraged the same collection of posts as \cite{turcan2019dreaddit}, with a specific focus on depression-related content. Two human annotators followed the DSM-5 criteria \cite{regier2013dsm} to categorize each post into four depression severity levels: minimum, mild, moderate, and severe. We used this dataset for post-level multiclass depression classification.
It includes 2,587, 290, 394, and 282 samples per class, with average token lengths of 105.3, 114.1, 111.6, and 109.2, respectively.  

\subsection{Baselines and Implementations}
To assess the effectiveness of the proposed model, we compare it against several representative methods that have demonstrated strong performance in uncertainty estimation. 
Dropout \citep{geifman2017selective,smith2018understanding}, as known as Monte Carlo (MC) dropout, is applied at test time to generate multiple stochastic predictions, providing an approximation of Bayesian uncertainty. 
Ensemble \citep{lakshminarayanan2017simple} trains multiple independently initialized deep models and aggregates their predictions, capturing model uncertainty through diversity among ensemble members. Uncertainty-Aware Attention (UA) \citep{heo2018uncertainty} models attention weights using a Gaussian distribution with learned mean and variance, allowing for more precise estimation of predictive uncertainty.
Evidential Deep Learning (EDL) \citep{sensoy2018evidential} models uncertainty by learning a Dirichlet distribution over class probabilities, allowing the network to distinguish between aleatoric and epistemic uncertainty.

The dataset was randomly divided into training (64\%), validation (16\%), and test (20\%) sets. Models were trained using the Adam optimizer \cite{kingma2014adam} with a learning rate of 1e-4, a mini-batch size of 12, and a maximum of 15 epochs, with early stopping applied to prevent overfitting.

For robust evaluation, 5-fold stratified cross-validation was conducted, and the mean performance metrics with their standard deviations were reported. Hyperparameters for all baseline models were tuned to ensure optimal performance.

\subsection{Evaluation}
We evaluate the proposed method from two complementary perspectives: predictive performance and uncertainty estimation. Predictive performance assesses the model’s ability to produce correct class labels, while uncertainty estimation examines whether the model can reliably indicate when its predictions are likely to be incorrect. Together, these evaluations provide a comprehensive view of both performance and reliability.

\subsubsection{Predictive Performance}
Predictive performance is evaluated using Accuracy (Acc), F1 score, Area Under the Receiver Operating Characteristic Curve (AUROC), and Area Under the Precision-Recall Curve (AUPRC). Accuracy measures the overall proportion of correct predictions but may be biased toward majority classes in imbalanced settings. To mitigate this limitation, we report the F1 score, which balances precision and recall and better reflects performance on minority classes \cite{powers2020evaluation}.

We further adopt AUROC to measure the model’s discriminative ability across different decision thresholds \cite{fawcett2006introduction}. As AUROC can be insensitive to class imbalance, we additionally report AUPRC, which focuses on precision-recall behavior and provides a more informative assessment when class distributions are skewed \cite{saito2015precision}. Reporting these metrics together enables a robust evaluation of predictive accuracy under imbalanced conditions.

\subsubsection{Uncertainty Estimation}
To evaluate uncertainty estimation, we measure how well predictive uncertainty correlates with misclassification using AUROC. Specifically, incorrect predictions are treated as positive samples, and the uncertainty score is used to discriminate between correct and incorrect outcomes. A higher AUROC indicates that the model assigns higher uncertainty to misclassified instances, reflecting more reliable uncertainty estimates.

This evaluation directly assesses the usefulness of uncertainty estimator for identifying unreliable predictions and is widely used in prior work on uncertainty-aware learning \cite{hendrycks2016baseline, lakshminarayanan2017simple}. As a threshold-independent metric, AUROC enables fair comparison across methods without additional calibration or tuning.

\section{Results}

\begin{table*}[!htbp]
\caption{Comparison of predictive performance between baseline methods and the proposed model across three datasets. The best results are shown in \textbf{bold}, and the second-best results are \underline{underlined}.
}
\label{tab:model_pred_comparisons}
\begin{tabular}{lllllllllll}
\hline
\multirow{2}{*}{Dataset}     & \multirow{2}{*}{Metric} & \multicolumn{2}{l}{Dropout}                                      & \multicolumn{2}{l}{Ensemble}                                       & \multicolumn{2}{l}{UA}                              & \multicolumn{2}{l}{EDL}                                      & Ours                              \\ \cline{3-11} 
                             &                         & Semantic                                 & Multi-view                      & Semantic                                 & Multi-view                     & Semantic                        & Multi-view                      & Semantic                        & Multi-view                               & Multi-view                               \\ \hline
\multirow{4}{*}{Dreaddit}    & Acc                     & 0.812{\tiny{$\pm$0.013}}          & 0.809{\tiny{$\pm$0.014}} & 0.816{\tiny{$\pm$0.017}}          & 0.811{\tiny{$\pm$0.029}} & 0.803{\tiny{$\pm$0.025}} & 0.811{\tiny{$\pm$0.011}} & 0.817{\tiny{$\pm$0.017}} & \underline{0.831{\tiny{$\pm$0.011}}}          & \textbf{0.835{\tiny{$\pm$0.017}}} \\
                             & F1                      & 0.819{\tiny{$\pm$0.014}}          & 0.818{\tiny{$\pm$0.021}} & 0.820{\tiny{$\pm$0.016}}          & 0.817{\tiny{$\pm$0.029}} & 0.809{\tiny{$\pm$0.029}} & 0.819{\tiny{$\pm$0.010}} & 0.828{\tiny{$\pm$0.014}} & \textbf{0.839{\tiny{$\pm$0.015}}} & \underline{0.837{\tiny{$\pm$0.023}}}          \\
                             & AUROC                   & 0.902{\tiny{$\pm$0.015}}          & 0.900{\tiny{$\pm$0.013}} & 0.905{\tiny{$\pm$0.011}}          & 0.894{\tiny{$\pm$0.025}} & 0.893{\tiny{$\pm$0.015}} & 0.893{\tiny{$\pm$0.013}} & 0.905{\tiny{$\pm$0.009}} & \underline{0.918{\tiny{$\pm$0.005}}}          & \textbf{0.928{\tiny{$\pm$0.007}}} \\
                             & AUPRC                   & 0.912{\tiny{$\pm$0.016}}          & 0.903{\tiny{$\pm$0.021}} & 0.913{\tiny{$\pm$0.012}}          & 0.902{\tiny{$\pm$0.026}} & 0.903{\tiny{$\pm$0.014}} & 0.895{\tiny{$\pm$0.028}} & 0.913{\tiny{$\pm$0.010}} & \underline{0.925{\tiny{$\pm$0.006}}}          & \textbf{0.937{\tiny{$\pm$0.006}}} \\ \hline
\multirow{4}{*}{SDCNL}       & Acc                     & \underline{0.725{\tiny{$\pm$0.023}}}          & 0.692{\tiny{$\pm$0.012}} & 0.726{\tiny{$\pm$0.013}}          & 0.696{\tiny{$\pm$0.022}} & 0.713{\tiny{$\pm$0.016}} & 0.693{\tiny{$\pm$0.031}} & 0.721{\tiny{$\pm$0.019}} & 0.657{\tiny{$\pm$0.015}}          & \textbf{0.731{\tiny{$\pm$0.021}}} \\
                             & F1                      & 0.734{\tiny{$\pm$0.036}}          & 0.718{\tiny{$\pm$0.011}} & \textbf{0.750{\tiny{$\pm$0.004}}} & 0.716{\tiny{$\pm$0.004}} & 0.737{\tiny{$\pm$0.031}} & 0.723{\tiny{$\pm$0.030}} & 0.736{\tiny{$\pm$0.032}} & 0.639{\tiny{$\pm$0.031}}          & \underline{0.743{\tiny{$\pm$0.024}}}          \\
                             & AUROC                   & \underline{0.801{\tiny{$\pm$0.015}}}          & 0.759{\tiny{$\pm$0.013}} & 0.799{\tiny{$\pm$0.012}}          & 0.758{\tiny{$\pm$0.024}} & 0.781{\tiny{$\pm$0.016}} & 0.753{\tiny{$\pm$0.025}} & 0.797{\tiny{$\pm$0.013}} & 0.729{\tiny{$\pm$0.011}}          & \textbf{0.810{\tiny{$\pm$0.015}}} \\
                             & AUPRC                   & \underline{0.804{\tiny{$\pm$0.014}}}          & 0.749{\tiny{$\pm$0.020}} & 0.803{\tiny{$\pm$0.011}}          & 0.761{\tiny{$\pm$0.026}} & 0.780{\tiny{$\pm$0.013}} & 0.733{\tiny{$\pm$0.026}} & 0.798{\tiny{$\pm$0.015}} & 0.746{\tiny{$\pm$0.022}}          & \textbf{0.805{\tiny{$\pm$0.010}}} \\ \hline
\multirow{4}{*}{DepSeverity} & Acc                     & \textbf{0.755{\tiny{$\pm$0.007}}} & 0.744{\tiny{$\pm$0.014}} & 0.747{\tiny{$\pm$0.009}}          & 0.738{\tiny{$\pm$0.018}} & 0.747{\tiny{$\pm$0.005}} & 0.746{\tiny{$\pm$0.022}} & 0.739{\tiny{$\pm$0.010}} & 0.731{\tiny{$\pm$0.013}}          & \underline{0.751{\tiny{$\pm$0.009}}}          \\
                             & F1                      & 0.416{\tiny{$\pm$0.037}}          & 0.387{\tiny{$\pm$0.087}} & \underline{0.432{\tiny{$\pm$0.045}}}          & 0.401{\tiny{$\pm$0.066}} & 0.379{\tiny{$\pm$0.025}} & 0.425{\tiny{$\pm$0.072}} & 0.327{\tiny{$\pm$0.026}} & 0.288{\tiny{$\pm$0.022}}          & \textbf{0.496{\tiny{$\pm$0.038}}} \\
                             & AUROC                   & \underline{0.841{\tiny{$\pm$0.013}}}          & 0.832{\tiny{$\pm$0.012}} & \textbf{0.849{\tiny{$\pm$0.012}}} & 0.835{\tiny{$\pm$0.008}} & 0.816{\tiny{$\pm$0.020}} & 0.822{\tiny{$\pm$0.027}} & 0.820{\tiny{$\pm$0.011}} & 0.792{\tiny{$\pm$0.020}}          & 0.835{\tiny{$\pm$0.009}}          \\
                             & AUPRC                   & 0.496{\tiny{$\pm$0.023}}          & 0.480{\tiny{$\pm$0.024}} & \underline{0.506{\tiny{$\pm$0.015}}}          & 0.502{\tiny{$\pm$0.025}} & 0.456{\tiny{$\pm$0.025}} & 0.476{\tiny{$\pm$0.050}} & 0.438{\tiny{$\pm$0.021}} & 0.405{\tiny{$\pm$0.016}}          & \textbf{0.517{\tiny{$\pm$0.033}}} \\ \hline
\end{tabular}

\end{table*}

\subsection{Performance Comparison with SOTAs}

In this section, we compare the proposed model with state-of-the-art (SOTAs) methods under both semantic-view and multi-view settings, focusing on predictive performance and uncertainty estimation.

\subsubsection{Predictive Performance}

Table \ref{tab:model_pred_comparisons} summarizes the predictive performance of different models on three datasets, evaluated using accuracy, F1 score, AUROC, and AUPRC. Overall, incorporating reasoning views in the baseline models does not consistently improve predictive performance.

In contrast, our proposed framework consistently outperformed the baseline models, demonstrating the effectiveness of the proposed multi-view integration strategy. On the Dreaddit dataset, our framework achieved the best results, with an accuracy of 0.853, an AUROC of 0.928, and an AUPRC of 0.937. Compared with the strongest baseline, this corresponded to improvements of approximately 0.48\% in accuracy, 1.09\% in AUROC, and 1.30\% in AUPRC.

On the SDCNL dataset, our framework again attained the highest performance by achieving an accuracy of 0.731, an AUROC of 0.810, and an AUPRC of 0.805. These results improved upon the best baseline by 0.83\% in accuracy, 1.12\% in AUROC, and 0.12\% in AUPRC.

For the DepSeverity dataset, our framework yielded the strongest performance in terms of F1 score and AUPRC, with values of 0.496 and 0.517, respectively. Compared to the best baseline model, this represented a substantial improvement of 14.81\% in F1 score and 2.17\% in AUPRC.

\begin{table}[!h]
\caption{Comparison of AUROC for predictive unceratinty and misclassification between baseline methods and the proposed model across three datasets. The best results are shown in \textbf{bold}, and the second-best results are \underline{underlined}.
}
\label{tab:model_uncertainty_comparisons}
\begin{tabular}{lllll}
\hline
\multicolumn{2}{l}{Model}         & Dreaddit                          & SDCNL                             & DepSeverity                       \\ \hline
\multirow{2}{*}{Dropout} & Semantic   & 0.534{\tiny{$\pm$0.038}}          & 0.464{\tiny{$\pm$0.029}}          & 0.509{\tiny{$\pm$0.059}}          \\
                     & Multi-view & 0.537{\tiny{$\pm$0.053}}          & 0.503{\tiny{$\pm$0.047}}          & 0.574{\tiny{$\pm$0.067}}          \\ \hline
\multirow{2}{*}{Ensemble}  & Semantic   & 0.576{\tiny{$\pm$0.035}}          & 0.498{\tiny{$\pm$0.026}}          & 0.423{\tiny{$\pm$0.060}}          \\
                     & Multi-view & 0.517{\tiny{$\pm$0.081}}          & 0.377{\tiny{$\pm$0.025}}          & 0.576{\tiny{$\pm$0.047}}          \\ \hline
\multirow{2}{*}{UA}  & Semantic   & 0.408{\tiny{$\pm$0.041}}          & 0.465{\tiny{$\pm$0.025}}          & 0.288{\tiny{$\pm$0.047}}          \\
                     & Multi-view & 0.414{\tiny{$\pm$0.038}}          & 0.452{\tiny{$\pm$0.051}}          & 0.196{\tiny{$\pm$0.022}}          \\ \hline
\multirow{2}{*}{EDL} & Semantic   & \underline{0.781{\tiny{$\pm$0.013}}}          & \underline{0.645{\tiny{$\pm$0.035}}}          & \textbf{0.852{\tiny{$\pm$0.009}}} \\
                     & Multi-view & 0.780{\tiny{$\pm$0.036}}          & 0.626{\tiny{$\pm$0.018}}          & 0.837{\tiny{$\pm$0.023}}          \\ \hline
Ours                 & Multi-view & \textbf{0.813{\tiny{$\pm$0.015}}} & \textbf{0.699{\tiny{$\pm$0.023}}} & \underline{0.846{\tiny{$\pm$0.009}}}          \\ \hline
\end{tabular}
\end{table}
\subsubsection{Uncertainty Estimation}

Table \ref{tab:model_uncertainty_comparisons} reports the performance of different models in uncertainty estimation across three datasets. On the Dreaddit dataset, our framework achieved the highest AUROC of 0.813, outperforming the best baseline by 4.10\%. Similarly, on the SDCNL dataset, our framework obtained the top AUROC of 0.699, yielding an improvement of 8.37\% over the strongest baseline.

For the baseline methods, uncertainty estimation performance was generally weaker. With the exception of EDL, most baseline models exhibit poor alignment between uncertainty scores and misclassification events.

\begin{figure*}[!h]
    \centering
    \includegraphics[width=1\linewidth]{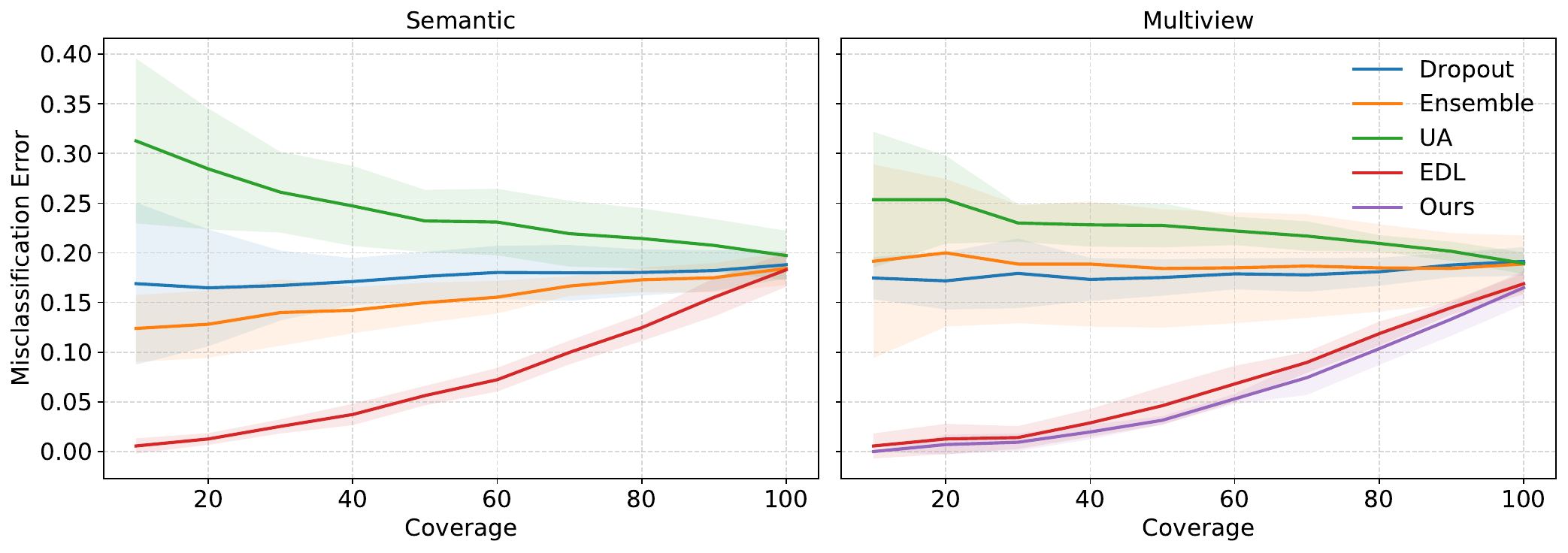}
    \caption{Risk coverage curve in \textbf{dreaddit} dataset.}
    \label{fig:risk_coverage_dreaddit}
\end{figure*}

\begin{figure*}[!h]
    \centering
    \includegraphics[width=1\linewidth]{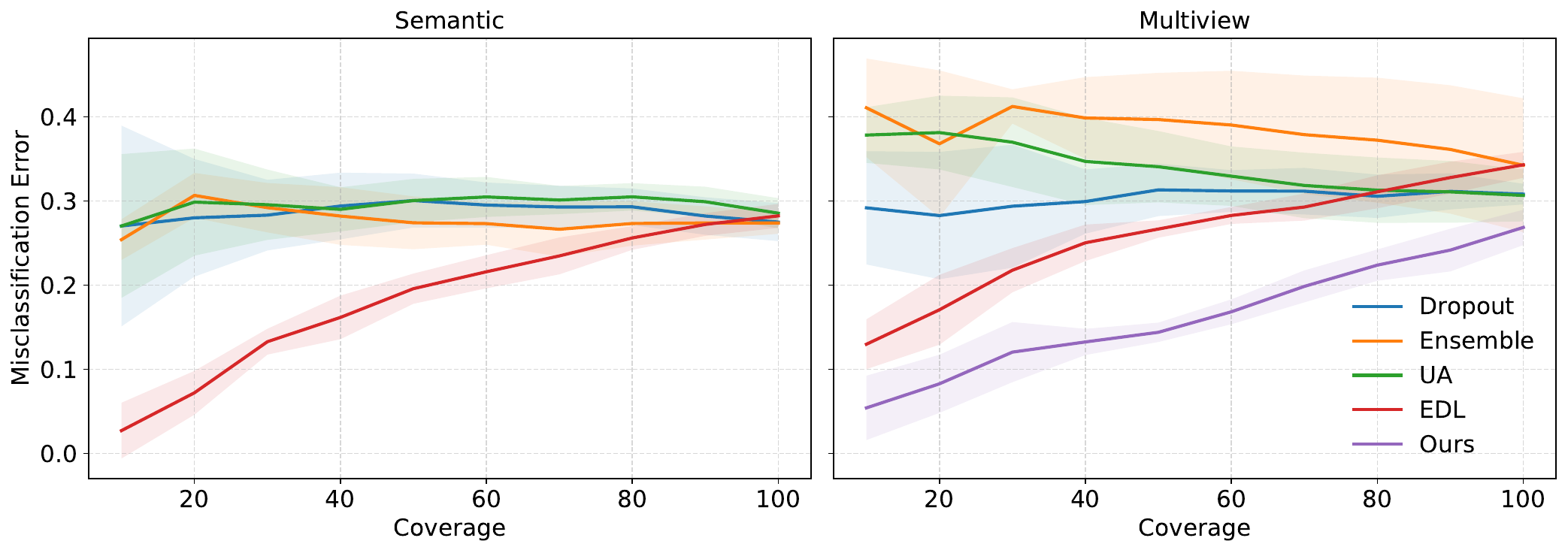}
    \caption{Risk coverage curve in \textbf{SDCNL} dataset.}
    \label{fig:risk_coverage_SDCNL}
\end{figure*}

\begin{figure*}[!h]
    \centering
    \includegraphics[width=1\linewidth]{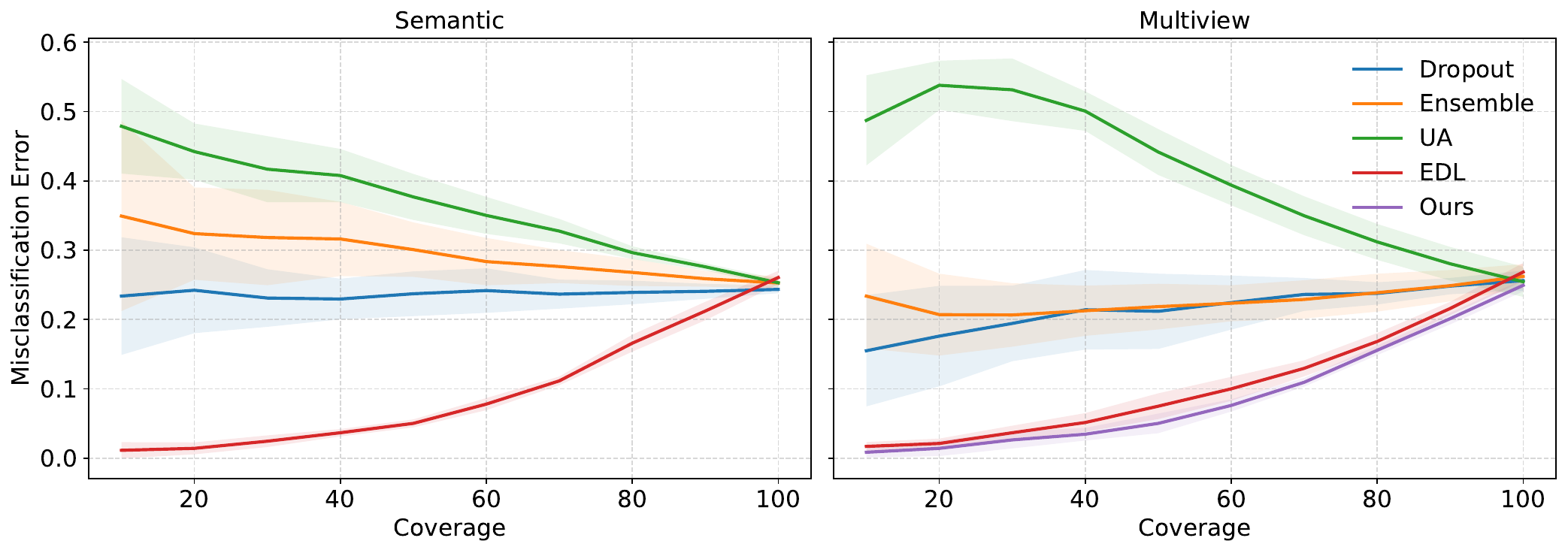}
    \caption{Risk coverage curve in \textbf{DepSeverity} dataset.}
    \label{fig:risk_coverage_DepSeverity}
\end{figure*}

To further examine model behavior, we analyzed the risk–coverage curves on the Dreaddit dataset in Figure \ref{fig:risk_coverage_dreaddit}, which depict how the misclassification error varies as the proportion of covered samples changes when samples were ranked by uncertainty from high to low. For evidence-based approaches, including EDL and our proposed method, the misclassification error consistently decreased as predictions were restricted to more confident samples. Similar observations on the other two datasets were reported in Figures \ref{fig:risk_coverage_SDCNL} and \ref{fig:risk_coverage_DepSeverity}.

\subsection{Performance Comparison on Different Views}

\begin{table}[!h]
\centering
\caption{Effect of removing individual views on predictive performance and uncertainty estimation in Dreaddit dataset.}
\label{tab:ablation}
\begin{tabular}{llll}
\hline
\multirow{2}{*}{Model} & \multicolumn{2}{l}{Predictive Performance}          & \makecell[l]{Uncertainty\\Estimation} \\ \cline{2-4} 
                       & AUROC                    & AUPRC                    & AUROC                                 \\ \hline
Complete Model         & 0.928{\tiny{$\pm$0.007}} & 0.937{\tiny{$\pm$0.006}} & 0.813{\tiny{$\pm$0.015}}              \\
- Semantic View        & 0.926{\tiny{$\pm$0.006}} & 0.933{\tiny{$\pm$0.007}} & 0.799{\tiny{$\pm$0.022}}              \\
- Symptom View         & 0.925{\tiny{$\pm$0.012}} & 0.933{\tiny{$\pm$0.013}} & 0.809{\tiny{$\pm$0.022}}              \\
- Emotion View         & 0.927{\tiny{$\pm$0.005}} & 0.935{\tiny{$\pm$0.006}} & 0.807{\tiny{$\pm$0.011}}              \\
- Cognitive View       & 0.926{\tiny{$\pm$0.005}} & 0.935{\tiny{$\pm$0.006}} & 0.796{\tiny{$\pm$0.018}}              \\ \hline
\end{tabular}
\end{table}

To examine the contribution of each view, we conducted an ablation study on the Dreaddit dataset by removing one view at a time while keeping the remaining components unchanged. Table~\ref{tab:ablation} reports the results in terms of predictive performance (AUROC and AUPRC) and uncertainty estimation (AUROC). The complete model achieves the highest predictive performance (AUROC: 0.928, AUPRC: 0.937) and the best uncertainty estimation (AUROC: 0.813).

Overall, the complete model achieves the highest predictive performance (AUROC: 0.928, AUPRC: 0.937) and the best uncertainty estimation (AUROC: 0.813). Removing any single view leads to slight decreases in predictive metrics. For uncertainty estimation, the drops are more noticeable, particularly when removing the semantic or cognitive view, with AUROC decreasing to 0.799 and 0.796, respectively.

\subsection{Robustness to Data Noise}

To assess robustness under realistic conditions, we introduced typographical noise into the input text. The noise was applied at the character level to preserve semantic content while simulating common typing errors in user-generated text. For each character, noise was injected with a fixed probability $p$, and one of four operations (deletion, insertion, substitution, or swap) was randomly applied. Insertion and substitution operations sampled characters from neighboring keys on a simplified QWERTY keyboard, reflecting realistic mistyping behavior. We evaluated robustness across multiple noise level by varying $p \in \{0.05,0.10,0.15,0.20,0.25\}$, generating test noisy datasets accordingly. Further implementation details can be found in Appendix \ref{appendix:noise}.

\begin{figure}[!h]
    \centering
    \includegraphics[width=1\linewidth]{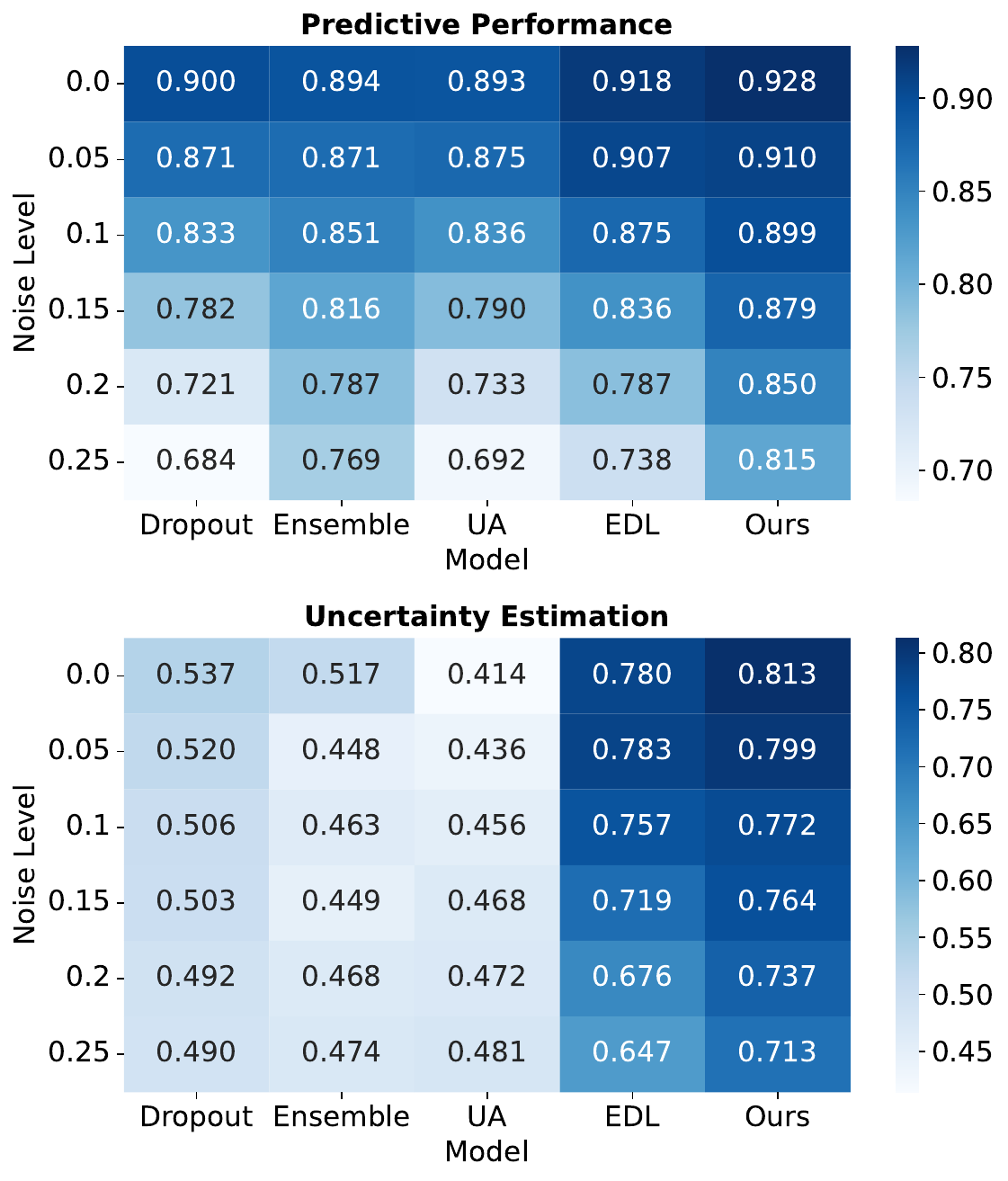}
    \caption{Heatmaps showing AUROC for predictive performance and uncertainty estimation across noise levels (rows) and models (columns) in Dreaddit dataset. Darker colors indicate higher performance, illustrating robustness to noise}
    \label{fig:data_noise}
\end{figure}

Figure \ref{fig:data_noise} presents heatmaps illustrating the AUROC for both predictive performance and uncertainty estimation across different noise levels.
As expected, all models experienced a decline in predictive accuracy as the noise level increased. However, our proposed method demonstrated strong robustness, maintaining relatively high AUROC even under substantial noise. 
For uncertainty estimation, our method exhibited a higher AUROC compared to all baselines. Notably, EDL model also showed competitive performance in uncertainty estimation, outperforming other standard baselines.

\section{Discussion}
In this section, we provide further discussion across several key findings: (1) the benefits of multi-view learning with evidential fusion, (2) the role of reliable uncertainty estimation in mental health prediction, (3) robustness to data quality, and (4) prediction interpretability.

\subsection{Benefits of Multi-View Learning with Evidential Fusion}

The results show that reasoning-awared multi-view learning does not always lead to better performance. Simply adding the reasoning views to baseline models does not consistently improve predictive accuracy and can even cause slight performance drops in some cases. This suggests that reasoning-based features may introduce noise or redundant information when their reliability is not explicitly modeled. One exception is the Dreaddit dataset, where multi-view fusion with EDL already performs well, indicating that evidence-based fusion can partly address this issue.

In contrast, our proposed framework consistently achieves the best predictive performance across all datasets. These results highlight the importance of evidential fusion when integrating different views. By treating each view as a source of evidence rather than giving all views equal weight, the framework can focus on informative signals while reducing the influence of noisy or unreliable inputs. This design is especially suitable for mental health prediction tasks, where the usefulness of different views can vary across datasets.

Beyond predictive performance, the benefits of evidence-based multi-view learning are more evident in uncertainty estimation. Our framework with evidential fusion achieves high AUROC values for uncertainty estimation on all datasets, showing a strong relationship between predictive uncertainty and misclassification events.

In comparison, most baseline methods show weak uncertainty estimation, with uncertainty scores that are poorly aligned with prediction errors. This limits their reliability in practical settings, particularly for high-stakes mental health applications.

Ablation results in Table~\ref{tab:ablation} further show that all views contribute to model performance, though their importance differs. The symptom view is most critical for prediction, as its removal leads to the largest drops in AUROC and AUPRC. For uncertainty estimation, semantic and cognitive views play a more important role, suggesting that higher-level contextual and reasoning information helps identify ambiguous or difficult cases.

Overall, these findings suggest that the main advantage of the proposed approach lies not only in improving predictive performance, but also in its ability to provide reliable uncertainty estimates through evidence-based multi-view fusion, which is crucial for risk-sensitive applications in mental health.

\subsection{Uncertainty Estimation Helps in Reliable Mental Health Prediction}

Incorporating the reasoning view through evidential fusion can help to improve the uncertainty estimation performance, as shown in Table \ref{tab:model_uncertainty_comparisons}. This suggests that although reasoning information may not always enhance predictive accuracy on their own, they provide complementary information when the uncertainty is explicitly modeled within an evidence-based framework.

\begin{figure}[!h]
    \centering
    \includegraphics[width=1\linewidth]{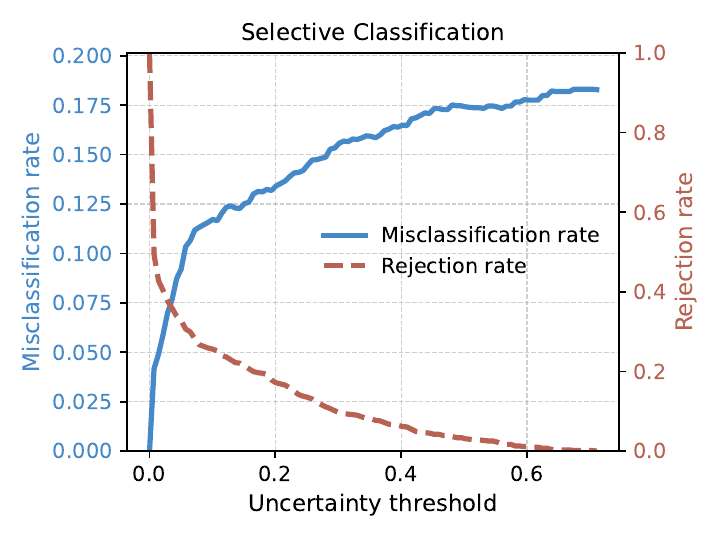}
    \caption{The curves of misclassification rate and rejection rate of our proposed model with respect to uncertainty threshold.}
    \label{fig:selective_classification}
\end{figure}

Figure \ref{fig:risk_coverage_dreaddit} indicates that our model and EDL can effectively identify high confident samples and make reliable selective predictions, which is crucial for mental health prediction. 
Figure \ref{fig:selective_classification} further illustrates this behavior by showing the misclassification rate and rejection rate of our model with respect to the uncertainty threshold. The rejection rate represents the proportion of samples deferred for human evaluation. As the uncertainty threshold decreased, the model achieved higher prediction accuracy by retaining only the most confident samples, while simultaneously rejecting a larger portion of samples. This highlights an inherent trade-off between misclassification rate and rejection rate when selecting an appropriate uncertainty threshold. A stricter threshold improves reliability but increases the workload for human experts, whereas a more relaxed threshold reduces rejections at the cost of higher prediction risk. Identifying a suitable balance is therefore critical for real-world deployment.

Such selective prediction behavior is particularly important for mental health applications, where incorrect predictions can have serious consequences. By prioritizing high-confidence predictions and deferring uncertain cases to human professionals, the proposed framework enhances reliability and supports safer deployment in high-stakes mental health settings.

\begin{figure*}[t]
    \centering
    \includegraphics[width=1\linewidth]{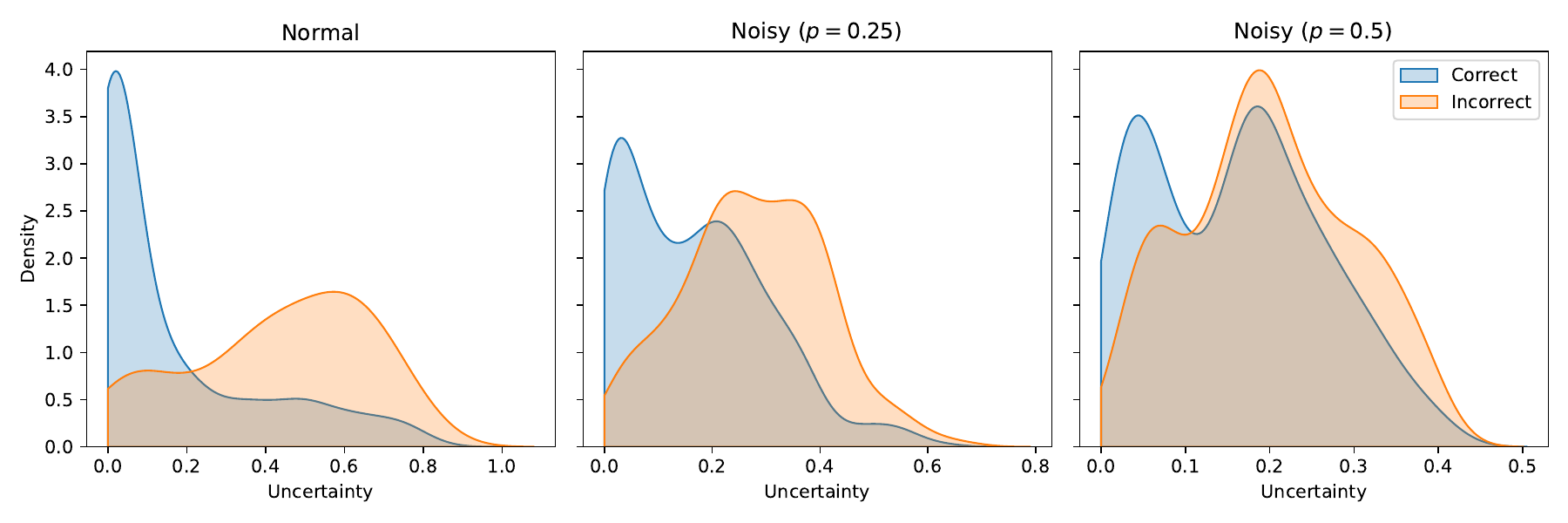}
    \caption{Kernel density estimates of predictive uncertainty conditioned on prediction correctness under original test data, noisy data ($p=0.25$), and noisy data ($p=0.5$).}
    \label{fig:uncertainty_density}
\end{figure*}

\subsection{Robustness to Data Quality}
The results in Figure \ref{fig:data_noise} indicate that, in addition to achieving the best predictive performance, our proposed model exhibited a stronger correlation between predicted uncertainty and actual misclassification rates. Among the baselines, the ensemble approach remained competitive in terms of predictive performance; however, its uncertainty estimates showed a weaker correlation with misclassification rates. Overall, these results demonstrate the advantage of our approach in providing both accurate predictions and trustworthy uncertainty estimation, which is essential for risk-sensitive applications with different levels of noise.

We further visualized the uncertainty density of our proposed model for correctly and incorrectly predicted samples under normal test data, noisy data ($p=0.25$), and noisy data ($p=0.5$), as shown in Figure \ref{fig:uncertainty_density}. For the normal test data, most correctly predicted samples were associated with low uncertainty, whereas incorrectly predicted samples exhibited higher uncertainty, indicating that the model was less confident when making incorrect predictions. After introducing noise, the test data became increasingly out of distribution. Under these conditions, the model continued to assign higher uncertainty to incorrectly predicted samples. However, the uncertainty of correctly predicted samples also increased, as the added noise degraded the quality of the input features and reduced the model’s confidence even when the final prediction remained correct.

\begin{figure*}
    \centering
    \includegraphics[width=1\linewidth]{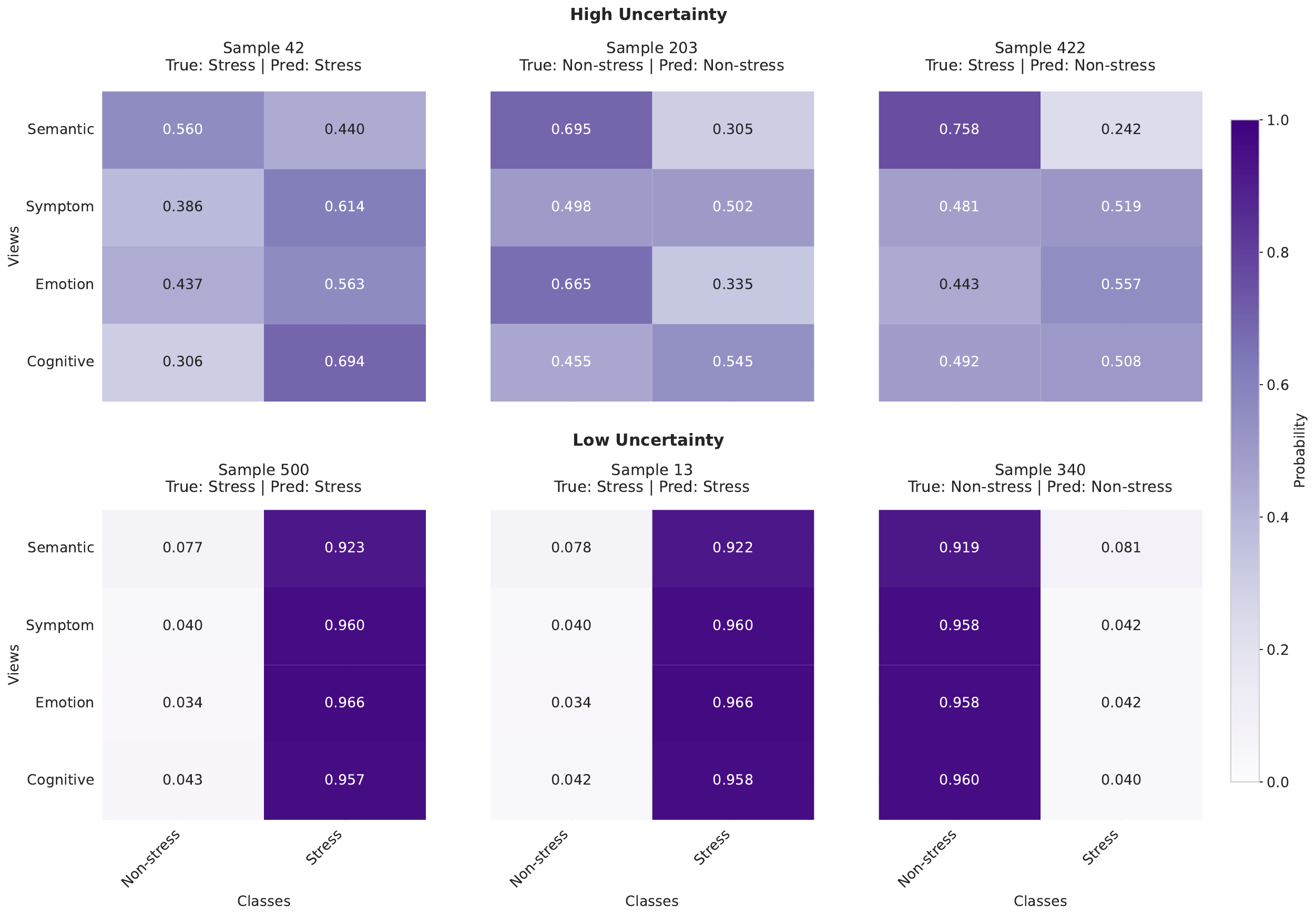}
    \caption{The probability distribution across different views for the samples with high uncertainty and low uncertainty.}
    \label{fig:case_study}
\end{figure*}

\begin{table*}[!h]
\centering
\caption{Examples of stress-related samples with high uncertainty and low uncertainty evaluated by our proposed model.}
\label{tab:qualitative}
\footnotesize
\renewcommand{\arraystretch}{0.95}

\textbf{High Uncertainty}

\vspace{0.3em}
\begin{tabular}{p{1cm}p{13.5cm}}
\toprule
Index & Sample \\
\midrule
42 &
I don't know if I'm able to do that, since I didn't tell the doctor the truth to begin with. I will say that my doctor very clearly didn't believe me at my follow up appointment, but my bf was in the waiting room. So I didn't admit to anything. I was teary eyed, however. From anyone's experience- is it worth it to go to the police?  \\ \hline

203 &
When I dare to open up to my friends, they tell me ‘Oh, my brother used to fight with me too’ and I feel diminished and dismissed. Since I left home at 21 and moved to the other side of the country I tried to talk up my family. I started from a place of recognising their vulnerability, justifying their actions by saying they were distracted by our financial woes and my dads early onset dementia. I told everyone they were amazing people. For the past two decades I have fought my body.    \\ \hline

422 &
Unfortunately, money is what me and her don't have now. My friends' help has dried up, and I don't blame them one bit for that. After five months, I've exhausted pretty much all the friends I can. Going into every little thing that has happened could fill a book. Maybe I \*WILL\* write a book about it someday, I don't know. \\
\bottomrule
\end{tabular}

\vspace{0.8em}

\textbf{Low Uncertainty}

\vspace{0.3em}
\begin{tabular}{p{1cm}p{13.5cm}}
\toprule
Index & Sample  \\
\midrule
500 &
I made a mistake as a result of being shaken up. Now she has complained to my boss but has falsely amplified my one mistake into me being absolutely negligent and incompetent. I found out about this 8 hours ago, had another flashback/panic attack, and my heart is still pounding. I've convinced myself I'm going to get fired. I know I need to go back to counseling. \\ \hline

13 & 
I'm worried I have a blood clot or something that gets aggravated when I wear them. I just want to be okay and to have a good time on this trip, but I've been so out of it, and I'm at my wit's end. Right now I'm lying down and I feel blood rushing all through my head, and bulging of blood vessels around my nose. I'm extremely sleep deprived and woozy but I'm scared to go to sleep and am in pain. I'm so scared.  \\ \hline

340 &
Hi everyone, I'm studying at the University of Maryland, and I'm working with a team on a product design project. My team and I have decided to focus on the idea of wellness with a particular focus on hospital/patient wellness. If you've ever spent any amount of time in a hospital, we want to hear from you! It's a short survey and would benefit us greatly in taking in feedback and redesigning our product. Thanks!  \\
\bottomrule
\end{tabular}

\end{table*}

\subsection{Prediction Interpretability} 
In this section, we examined the interpretability of our multi-view framework through case analysis under different levels of uncertainty. Figure~\ref{fig:case_study} shows the predicted probability distributions for each view in samples with high and low uncertainty, while Table~\ref{tab:qualitative} provides the corresponding text examples.

For samples with high uncertainty, the semantic view produces relatively confident and consistent predictions, whereas the reasoning views (symptom, emotion, and cognitive) exhibit more dispersed and ambiguous probability distributions. As shown in Table~\ref{tab:qualitative}, these samples often involve complex narratives, implicit distress, and contextual ambiguity (e.g., interpersonal conflicts, long-term family issues, or financial hardship). While the overall semantic content conveys stress-related meaning at a coarse level, explicit signals required by the reasoning views such as clearly stated symptoms, identifiable emotional states, or structured cognitive patterns are intertwined with background context. This makes fine-grained reasoning-based interpretation more difficult and leads to higher predictive uncertainty in the reasoning views.
However, in sample 42, the semantic view made the wrong prediction, while the reasoning views, especially the cognitive view, aligned better with the true label. The text in Table~\ref{tab:qualitative} shows patterns of cognitive distortions such as mind-reading, catastrophizing, and personalization. For example, the individual assumes the doctor did not believe them (mind-reading), doubts their own ability to act (catastrophizing), and takes responsibility for the situation (personalization). These reasoning cues, captured by the cognitive and emotion views, provide structured signals that are more directly linked to stress-related indicators, explaining why the reasoning views produced the correct predictions in this case despite the semantic ambiguity.

In contrast, samples with low uncertainty showed a different pattern. 
As shown in Figure~\ref{fig:case_study}, the reasoning views produced sharper and more confident distributions that matched the semantic view. These samples provided clear and concrete indicators, such as physiological symptoms (panic attacks, heart pounding), strong emotions (fear, anxiety), and identifiable cognitive patterns (catastrophic or health-related worry). These explicit cues allow reasoning views to extract structured information more easily, leading to lower uncertainty and consistent predictions across views.

Overall, this case studies demonstrate that the proposed multi-view framework captures complementary interpretability under varying uncertainty levels. When reasoning cues are implicit or diffuse, the semantic view provides a more reliable signal, whereas in cases with explicit symptoms, emotions and cognitive patterns, reasoning-based views contribute more decisively. This behavior aligns with human intuition and highlights the advantage of integrating semantic understanding with structured reasoning for trustworthy mental health prediction.

\section{Conclusion}
In this work, we propose a reasoning-aware multi-view learning framework for automated mental health prediction. By modeling semantic information from encoder-only models with reasoning information from decoder-only models via Subjective Logic, and integrating them through an evidential fusion strategy, our approach improved both prediction accuracy and reliability. It explicitly models uncertainty, allowing the model to handle noisy data more robustly. Experiments on multiple benchmark datasets shows that our method not only achieved strong predictive performance but also provided trustworthy uncertainty estimates and interpretable reasoning information. These results suggest that our framework is suitable for risk-sensitive mental health applications where both accuracy and reliability are crucial.

\section*{Acknowledgments}
This work is partially supported by the National Medical Research Council via the HEALTHY AND MEANINGFUL LONGEVITY - COGNITION GRANT with project ID NICCOG2024-0028 and the AI for Public Health Program at Saw Swee Hock School of Public Health, National University of Singapore.


%

\bibliographystyle{IEEEtran}
\bibliography{references}

\end{document}